\documentclass[sigconf]{acmart}
\renewcommand\footnotetextcopyrightpermission[1]{} 
\usepackage[utf8]{inputenc}
\usepackage{hyperref}
\usepackage{graphicx}
\usepackage[normalem]{ulem} 
\usepackage{booktabs}
\usepackage{balance}
\usepackage{multirow}
\usepackage{subcaption}

\renewcommand{\paragraph}[1]{\smallskip\noindent\textbf{#1.\mbox{\ \ }}}
\newcommand{\sr}[1]{\textit{\textcolor{blue}{SR: #1}}}
\newcommand{\yc}[1]{\textit{\textcolor{orange}{YC: #1}}}
\newcommand{\ex}[1]{{\small \texttt{\small #1}}}

\newcommand{\dice}{\textsc{Dice}}

\usepackage{tikz}
\usepackage{pgfplots}
\pgfplotsset{compat=newest}
\usepgfplotslibrary{groupplots}
\usepgfplotslibrary{dateplot}

\usepackage{xcolor}
\usepackage{colortbl}
\definecolor{custom-gray}{rgb}{0.85, 0.85, 0.85}

\newcommand{\squishlist}{ 

 \begin{list}{$\bullet$}
  { \setlength{\itemsep}{0pt}
     \setlength{\parsep}{1pt}
     \setlength{\topsep}{1pt}
     \setlength{\partopsep}{0pt}
     \setlength{\leftmargin}{1.5em}
     \setlength{\labelwidth}{1em}
     \setlength{\labelsep}{0.5em} } }
 \newcommand{\squishend}{\end{list}}

\title{Joint Reasoning for Multi-Faceted
Commonsense Knowledge}

\author{Yohan Chalier}
 \affiliation{
     \institution{Télécom ParisTech}
 }
 \email{yohan@chalier.fr}

 \author{Simon Razniewski}
 \affiliation{
     \institution{Max Planck Institute for Informatics}
 }
 \email{srazniew@mpi-inf.mpg.de}

 \author{Gerhard Weikum}
 \affiliation{
     \institution{Max Planck Institute for Informatics}
 }
 \email{weikum@mpi-inf.mpg.de}

\begin{document}

\begin{abstract}
Commonsense knowledge (CSK) supports
 a variety of AI applications, from visual understanding to chatbots.
 Prior works on acquiring CSK, such as ConceptNet, 
 have compiled statements that associate
 concepts, like everyday objects or activities,
 with properties that hold for most or some instances of the concept.
 Each concept is treated in isolation from other concepts, and the only quantitative measure
 (or ranking) of properties is a confidence score that the statement is valid.

 This paper aims to overcome these limitations by introducing 
 a multi-faceted model of CSK statements and 
 methods for joint reasoning over
 sets of inter-related statements.
 %
  Our model captures four different dimensions of CSK statements:
  plausibility, typicality, remarkability and salience, with scoring and ranking along each dimension. 
  For example, hyenas drinking water is typical but not salient, whereas hyenas eating carcasses
  is salient.
  %
  For reasoning 
  and ranking, 
  we develop a method with soft constraints, 
  to couple the inference over concepts that are related in 
  in a taxonomic hierarchy.
  The reasoning is cast into an integer linear programming (ILP), and we leverage the theory of reduction costs of a relaxed LP to compute informative rankings.
%
    This methodology is applied to 
    several large CSK collections. Our evaluation shows that we can consolidate these inputs into much cleaner and more expressive knowledge. Results are available at \url{https://dice.mpi-inf.mpg.de}.
\end{abstract}

\maketitle

\section{Introduction}

\paragraph{Motivation and problem}
Commonsense knowledge (CSK) is a
potentially important asset 
towards building versatile AI applications, such as 
visual understanding for 
describing images 
(e.g., \cite{DBLP:journals/ijcv/AgrawalLAMZPB17,DBLP:journals/pami/KarpathyF17,Shuster_2019_CVPR})
or conversational agents
like chatbots 
(e.g., \cite{DBLP:conf/aaai/YoungCCZBH18,DBLP:conf/acl/RajaniMXS19,DBLP:conf/acl/YangLLLS19}).
In delineation from encyclopedic knowledge on entities like Trump, Paris, or FC Liverpool, CSK refers to properties, 
traits and relations of everyday concepts, such as elephants, coffee mugs or school buses. 
For example, when seeing scenes of an elephant
juggling a few coffee mugs with its trunk, or with 
school kids pushing  an elephant into a bus,
an AI agent with CSK should realize the
absurdity of these scenes and should generate
funny comments for image description or in a 
conversation.

Encyclopedic knowledge bases (KBs)
received much attention, with projects such as
DBpedia~\cite{Auer:ISWC2007},
Wikidata~\cite{vrandevcic2014wikidata}, Yago~\cite{suchanek2007yago} or NELL~\cite{nell} 
and large knowledge graphs at
Amazon, Baidu, Google, Microsoft etc.
supporting
entity-centric search and other
services \cite{Noy:CACM2019}.
In contrast, approaches to acquire CSK
have been few and limited. 
Projects like ConceptNet~\cite{conceptnet}, WebChild~\cite{webchild}, TupleKB~\cite{TupleKB} and Quasimodo~\cite{quasimodo} 
have compiled
millions of 
{\em concept: property} (or subject-predicate-object)
statements, but still suffer from sparsity and noise. 
For instance, ConceptNet has only a single non-taxonomic/non-lexical statement about hyenas, namely, 
{\em hyenas: laugh a lot}
\footnote{\url{http://conceptnet.io/c/en/hyena}}, 
and
WebChild lists overly general and contradictory 
properties such as \emph{small, large, demonic} and \emph{fair}  for hyenas\footnote{\url{https://gate.d5.mpi-inf.mpg.de/webchild2/?x=hyena\%23n\%231}}. 
The reason for these shortcomings is that 
such mundane properties that are
obvious
to every human are
rarely expressed explicitly in text or speech, and visual content would require CSK first to extract
these properties. Therefore, 
machine-learning methods
for encyclopedic knowledge acquisition
do not work robustly for CSK.


Another limitation of existing CSK collections
is that they organize statements in a flat, 
one-dimensional manner, and solely rank by
confidence scores. 
There is no information about whether
a property holds for all or for some of
the instances of a concept, and there
is no awareness of which properties are
typical and which ones are salient
from a human perspective.
For example, the statement that hyenas
drink milk (as all mammals when they are cubs)
is valid, but it is not typical.
Hyenas eating meat is typical, but it is
not salient in the sense that humans 
would spontaneously name this as a key
characteristic of hyenas.
In contrast, hyenas eating carcasses
is remarkable as it sets hyenas apart
from other African predators (like lions
or leopards), and many humans would list
this as a salient property.
Prior works on CSK missed out
on these refined and expressive
dimensions. 

The problem addressed in this work is
to overcome these limitations and advance
CSK collections to a more expressive
stage of multi-faceted knowledge.


\paragraph{Approach and contribution}
This paper presents \dice{} (\underline{Di}verse \underline{C}ommonsense Knowl\underline{e}dge), 
a reasoning-based method 
for deriving 
refined and expressive
commonsense knowledge from existing CSK collections.
 \dice{} is based on two novel ideas:
\squishlist
\item To capture the refined semantics of CSK statements, we introduce four facets of concept properties:
\squishlist
\item {\em Plausibility} indicates whether a statement makes sense at all (like the established but overloaded 
notion of confidence scores).
\item {\em Typicality} indicates whether a property holds for most instances of a concept (e.g., not only for cubs).
\item {\em Remarkability} expresses that a property stands out by distinguishing the concept
from closely related concepts (like siblings in a taxonomy).
\item {\em Saliency} reflects that a property is characteristic for the concept,
in the sense that most humans would spontaneously list it in association with the concept.
\squishend
\item  We identify inter-related concepts by their 
neighborhoods in a concept hierarchy or via word-level embeddings,
and devised a set of weighted soft constraints that allows us to jointly reason over the four dimensions
for sets of candidate statements. We cast this approach into an integer linear program (ILP), and
harness the theory of reduced cost (aka. opportunity cost) \cite{bertsimas1997introduction} for LP relaxations in order
to compute quantitative rankings for each of the four facets.
\squishend
%

As an example, consider the concepts {\em lions}, {\em leopards}, {\em cheetahs} and {\em hyenas}.
The first three are coupled by being taxonomic siblings under their hypernym {\em big cats},
and the last one is highly related by being another predator in the African savannah
with high relatedness in word-level embedding spaces (e.g., word2vec or Glove).
Our constraint system includes 
logical clauses such as 
\squishlist
\item[ ] $\text{Plausible}(s_1, p) \land  \text{Related}(s_1, s_2) \land \lnot \text{Plausible}(s_2, p) \land \dots$
\item[ ] $\Rightarrow \text{Remarkable}(s_1, p)$
\squishend
where $\dots$ refers to enumerating all siblings of $s_1$, or highly related concepts.
The constraint itself is weighted by the degree of relatedness; so it is a soft constraint
that does allow exceptions.
This way we can infer that remarkable (and also salient) statements include 
{\em lions: live in prides}, {\em leopards: climb trees}, {\em cheetahs: run fast} and {\em hyenas: eat carcasses}.

The paper's salient contributions are:
\squishlist
    \item We introduce a multi-faceted model for CSK 
    statements, 
    comprising the dimensions of 
    plausibility, typicality, remarkability and saliency.
    \item We model the coupling of these dimensions by a soft constraint system, and devise effective and scalable
    techniques for joint reasoning over noisy candidate statements, 
    \item Experiments, with inputs from large CSK collections, ConceptNet, TupleKB and Quasimodo, and with human judgements,
    show that \dice{} achieves high precision for its multi-faceted output.
    The resulting commonsense knowledge bases contain more than 1.6m statements about 74k concepts,
    and will be made publicly available.
\squishend


\section{Related Work}


\paragraph{Manually compiled CSK} In 1985, Douglas Lenat started the Cyc~\cite{lenat1995cyc} project, with the goal of compiling a comprehensive machine-readable collection of human knowledge
into logical assertions.
The project comprised both encyclopedic and commonsense knowledge. 
The parallel WordNet project \cite{wordnet} organized word senses into lexical relations 
like synonymy, antonymy, and hypernymy/hyponymy (i.e., subsumption). 
The latter can serve as a taxonomic backbone for CSK, but there are also more recent alternatives
such as WebIsALOD \cite{HertlingPaulheim:ISWC2017} derived from Web contents.
ConceptNet extended the Cyc and WordNet approaches 
by collecting CSK triples from crowdworkers, for about 20 high-level properties~\cite{conceptnet}.
It is the state of the art for CSK.
The most popular knowledge base today, Wikidata~\cite{vrandevcic2014wikidata}, contains both
encyclopedic knowledge about notable entities and some CSK cast into RDF triples.
However, the focus is on individual entities, and CSK is very sparse.
Most recently, ATOMIC~\cite{atomic} is another crowdsourcing project compiling knowledge about
human activities; relative to ConceptNet it is more refined but fairly sparse.

\paragraph{Web-extracted CSK} Although handcrafted CSK collections have reached impressive sizes, 
the reliance on human inputs limits their scale and scope.
Automatic information extraction (IE) from Web contents can potentially achieve much higher coverage. 
Compared to general IE, 
extracting CSK is still an underexplored field.
The WebChild project \cite{webchild,Tandon:ACL2017} extracted more than 10 million statements of
plausible object properties from books and image tags. 
However, its rationale was to capture each and every property that holds for some instances of a concept;
consequently, it has a massive tail of noisy, puzzling or invalid statements.
%
TupleKB~\cite{TupleKB} from the AI2 Lab's Mosaic project
is a more focused approach to automatic CSK acquisition. It contains ca. 280k statements,
specifically for 8th-grade elementary science to support work on a multiple-choice school exam
challenge
\cite{DBLP:journals/cacm/SchoenickCTTE17}.
It builds on similar sources as WebChild, but prioritizes precision over recall by various
cleaning steps incl. a supervised scoring model.
%
Quasimodo~\cite{quasimodo} is a recent CSK collection, built by extraction from QA forums and web query logs, with about 4.6 million statements. Although it combines multiple cues into a regression-based
corroboration model for ranking and aims to identify salient statements, the model 
merely learns a single-dimensional notion of confidence.
%
Common to all these projects is that their quantitative assessment of CSK statements
is focused on a single dimension of confidence or plausibility. There is no awareness of
other facets like typicality, remarkability and saliency.

\paragraph{Latent representations} Latent models have had great impact on natural language
processing, with word embeddings like word2vec \cite{mikolov2013distributed}, GloVe~\cite{pennington2014glove} and BERT~\cite{devlin2019bert} capturing signals from huge text corpora.
These embeddings implicitly contain some kind of CSK by the relatedness of word-level or phrase-level vectors or
more advanced representation.
For example, 
the typical habitats for camels can be predicted to be deserts, based on the latent representations.
Embeddings have been leveraged for tasks like commonsense question answering \cite{commonsenseqa}
and knowledge base completion (e.g., \cite{bosselut2019comet}). 
However, the latent nature of these models makes it difficult to interpret what specific knowledge
is at work and explain this to the human user. Moreover, they typically involve a complete
end-to-end training cycle for each and every use case. Explicit CSK collections are much
better interpretable and more easily re-usable for new applications.

\paragraph{Joint reasoning}
Consolidating statements from automatic IE is an important part of KB construction, and several frameworks have been pursued for encyclopedic knowledge, including 
probabilistic graphical models of different kinds 
(e.g., \cite{DBLP:series/synthesis/2009Domingos,nell,DBLP:journals/pvldb/WickMM10,pujara2013knowledge,shin2015incremental},
constraint-based reasoning (e.g., \cite{suchanek2009sofie,DBLP:conf/emnlp/SrikumarR11}), and more.
All these methods solve optimization problems to accept or reject uncertain candidate statements
with specified or learned constraints so as maximize a combination of statistical evidence 
and satisfaction of soft constraints. 

%
\paragraph{Knowledge representation}
Current CSKBs merely use a single score that represents the frequency of or 
confidence in a binary-relation statement.
Beyond binary relations, epistemic 
logics would be able to express refined modalities such as possibly and necessarily. 
Temporal logics can model whether statements are valid always, eventually
or sometimes
\cite{ozaki2018happy}, and spatial data models can capture location information about entities and events~\cite{abraham1999survey}. 
The need to contextualize binary relations has been noted in encyclopedic KBs.
Yago introduced the notion of SPOTLX tuples to capture time, location and
textual dimensions~\cite{yago2,yahya2016relationship}, DBpedia used reification to store 
provenance information~\cite{hellmann2009dbpedia}, and Wikidata comes with a range of temporal, spatial, and other contextual qualifiers~\cite{patelcontextualization}. 
For CSKBs this level of refinement has not been considered yet.
In KG embeddings, Chen et al.\ studied models for retaining graded truth values, termed confidence, instead of binary truth values, in inputs and outputs of embedding models~\cite{chen2019embedding}. However, this is limited to a single
dimension, and does not capture the different facets addressed in this paper.

\section{Multi-Faceted CSK Model}


We consider simple CSK statements
of the form $(s,p)$, 
where $s$ is a concept and $p$ is a property of this concept.
To be in line with established terminology, we refer to $s$ as the
subject of the statement. Typically, $s$ is a crisp noun, such as {\em hyenas},
while $p$ can take any multi-word verb or noun phrase, such as
{\em laugh a lot} or {\em (are) African predators}.

Unlike prior works, we do not adopt the usual subject-predicate-object triple model.
We do not distinguish between predicates and objects for 
two reasons:\\
(i) The split between predicate and object is often arbitrary. For example, for
$lions: live~in~prides$, we could either consider {\em live} or {\em live in}
as predicate and the rest as object, or we could view {\em live in prides} as a predicate
without any object. \\
(ii) Unlike encyclopedic KBs where a common set of predicates can be standardized
(e.g., {\em date of birth}, {\em country of citizenship}, {\em award received}),
CSK is so diverse that it is virtually impossible to agree on predicate names.
For example, we may want to capture both {\em prey on antelopes} and {\em hunt and kill antelopes},
which are highly related but not quite the same.
Projects like ConceptNet and WebChild have organized CSK with a fixed set of pre-specified
predicates, but these are merely around 20, and, when discounting taxonomic (e.g., type of)
and lexical (e.g., synonyms, related terms) relations, boil down to a few basic predicates:
{\em used for}, {\em capable of}, {\em location} and {\em part of} (plus a generic kind of
{\em has property}).\\

We summarize important notation in Table~\ref{tab:notation}.

\begin{table}
\centering
\begin{tabular}{lp{5.5cm}}
\toprule
Term & Meaning \\ \midrule
CSK statement & pair $(s,p)$ of subject $s$ (concept) and property $p$ (textual phrase)\\
CSK dimensions & plausibility, typicality, remarkability, saliency\\
Soft constraints & Relationships between dimensions of a statement and/or taxonomically related concepts\\
Taxonomy & Noisy \textit{is-a} organization of subject concepts\\
Clause & A grounding, i.e., concrete instantiation of a rule\\
$\omega_r,\omega_s,\omega_e$ & Parameters for weighing clauses\\
Cues & Input signals for estimating prior scores \\ 
Prior scores & Initial estimates of dimension values for a statement (i.e., before reasoning), denoted as $\pi, \tau, \rho, \sigma$ and computed from cues via regression \\
\bottomrule
\end{tabular}
\caption{Important notation.}
\label{tab:notation}
\end{table}

\subsection{CSK Dimensions}

We organize concept-property pairs along four dimensions:
plausibility~\cite{webchild,mishra2017domain}, typicality~\cite{conceptnet}, remarkability (information theory) and saliency~\cite{quasimodo}. These are meta-properties;
so each $(s,p)$ pair can have any of these labels and multiple labels are possible. 
For each statement and dimension label, we compute a score and can thus rank
statements for a concept by their plausibility, typicality, remarkability or saliency.

\squishlist
   \item \emph{Plausibility:} 
   Is the property valid at least for some instances of the concept, for at least some
   spatial, temporal or socio-cultural contexts? For example, lions drink milk
   at some time in their lives, and some lions attack humans.
    \item \emph{Typicality:} 
Does the property hold for most (or ideally all) instances of the concept, for most contexts?
For example, most lions eat meat, regardless of whether they live in Africa or in a zoo.
    \item \emph{Remarkability:} 
    What are specific properties of a concept that sets the concept apart from highly related concepts,
    like taxonomic generalizations (hypernyms in a concept hierarchy)?
    For example, lions live in prides but not other big cats do this, and
    hyenas eat carcasses but hardly any other African predator does this.
     \item \emph{Saliency:} 
    When humans are asked about a concept, such as {\em lions}, {\em bicycles} or {\em rap songs},
    would a property be listed among the concept's most notable traits, by most people?
    For example, lions hunt in packs, bicycles have two wheels, rap songs have interesting lyrics and beat
    (but no real melody).
\squishend


\paragraph{Examples}
Refining CSK by the four dimensions is useful for various application areas, including
language understanding for chatbots, as illustrated by the following examples:
\begin{enumerate}
    \item Plausibility helps to avoid blunders by detecting absurd statements, or to trigger irony.
    For example, a user utterance such as ``When too many people shot selfies with him, 
    the lion king in the zoo told them to go home'' should lead to a funny reply by the chatbot
    (as lions do not speak).
        \item Typicality helps a chatbot to infer missing context. 
        For example, when the human talks about ``a documentary which showed the feeding frenzy
        of a pack of hyenas'', the chatbot could ask ``what kind of carcass did they feed on?''
    \item Remarkability can be an important signal when the chatbot needs to infer which concept the
    human is talking about. For example, a user utterance ``In the zoo, the kids where fascinated 
    by a spotted dog that was laughing at them'' could lead to chatbot response like
    ``So they like the hyenas. Did you see an entire pack?''
    \item Saliency enables the chatbot to infer important properties when a certain concept is
    the topic of a conversation. For example, when talking about lions in the zoo, the bot
    could proactively ask ``Did you hear the lion roar?'', or ``How many lionesses were in the
    lion king's harem?''
\end{enumerate}

\section{Joint Reasoning}

\paragraph{Overview}
For reasoning over sets of CSK statements,
we start with a CSK collection, like
ConceptNet, TupleKB or Quasimodo.
These are in triple form with crisp subjects
but potentially noisy phrases as
predicates and objects. We interpret each
subject as a concept and concatenate the
predicate and object into a property.
Inter-related subsets of statements
are identified by locating concepts
in a large taxonomy and grouping
siblings and their hypernymy parents
together. These groups may overlap.
For this purpose we use the
WebIsALOD taxonomy \cite{HertlingPaulheim:ISWC2017},
as it has very good coverage of
concepts and captures everyday vocabulary.

Based on the taxonomy, we also generate additional
candidate statements for sub- or super-concepts,
as we assume that many properties are inherited between
parent and child. We use rule-based templates 
for this expansion of the CSK collection
(e.g., as lions are predators, big cats and also
tigers, leopards etc. are predators as well).
This mitigates the sparseness in the observation space.
Note that, without the reasoning, this would be
a high-risk step as it includes many invalid statements
(e.g., lions live in prides, but big cats in general do not).
Reasoning will prune out most of the invalid candidates, though.

For joint reasoning over the statements for
the concepts of a group, we interpret the rule-based templates as soft constraints,
with appropriate weights.

For setting weights in a meaningful way,
we leverage prior scores
that the initial CSK statements come with
(e.g., confidence scores from ConceptNet),
and additional statistics from
large corpora, most notably word-level
embeddings like 
word2vec.

In this section, we develop the
logical representation and the joint reasoning method,
assuming that we have weights for
statements and for the grounded
instantiations of the constraints.
Subsequently, Section \ref{sec:priorscores}
presents
techniques for obtaining statistical priors
for setting the weights.




\subsection{Coupling of CSK Dimensions}

Let $\mathcal{S}$ denote the set of subjects and $\mathcal{P}$ the properties. 
The inter-dependencies between the four CSK dimensions are expressed
by the following
logical constraints.

\noindent{\bf Concept-dimension dependencies:}
$\forall (s, p)\in\mathcal{S}\times\mathcal{P}$
\begin{align}
    \text{Typical}(s, p) & \Rightarrow  \text{Plausible}(s, p) \label{eq:typical_implies_plausible}\\
    \text{Salient}(s, p) & \Rightarrow \text{Plausible}(s, p) \\
    \text{Typical}(s, p) \wedge \text{Remarkable}(s, p) & \Rightarrow \text{Salient}(s, p)
\end{align}
These clauses capture the intuition behind the four facets.

\noindent{\bf Parent-child dependencies:}
$\forall (s_1, p)\in\mathcal{S}\times\mathcal{P}, \forall s_2\in \text{children}(s_1)$ 
\begin{align}
    \text{Plausible}(s_1, p) & \Rightarrow \text{Plausible}(s_2, p) \label{eq:plausibility_inheritance_constraint} \\
    \text{Typical}(s_1, p) & \Rightarrow \text{Typical}(s_2, p) \\
    \text{Typical}(s_2, p) & \Rightarrow \text{Plausible}(s_1, p) \label{eq:plausibility_inference_constraint}\\
    \text{Remarkable}(s_1, p) & \Rightarrow \neg \text{Remarkable}(s_2, p) \\
    \text{Typical}(s_1, p) & \Rightarrow \neg \text{Remarkable}(s_2, p) \label{eq:typical_prevents_remarkable_children} \\
    \neg \text{Plausible}(s_1, p) \wedge \text{Plausible}(s_2, p) & \Rightarrow \text{Remarkable}(s_2, p) \\
    \left(\forall s_2 \in \text{children}(s_1) \>\> \text{Typical}(s_2, p) \right) & \Rightarrow \text{Typical}(s_1, p)
\end{align}
%
These dependencies state how properties are inherited between a parent concept and its children in a taxonomic hierarchy. For example, if a property is typical for the parent and thus for all its children, it is not remarkable for any child as it does not set any child apart from its siblings.

\noindent{\bf Sibling dependencies:}
$\forall (s_1, p)\in\mathcal{S}\times\mathcal{P}, \forall s_2\in \text{siblings}(s_1)$ 
\begin{align}
    \text{Remarkable}(s_1, p) & \Rightarrow \neg \text{Remarkable}(s_2, p) \label{eq:remarkability_incompatibility_siblings_contraint} \\
    \text{Typical}(s_1, p) & \Rightarrow \neg \text{Remarkable}(s_2, p) \label{eq:typical_siblings_constraint} \\
    \neg \text{Plausible}(s_1, p) \wedge \text{Plausible}(s_2, p) & \Rightarrow \text{Remarkable}(s_2, p) \label{eq:not_plausible_siblings_contraints}
\end{align}
These dependencies state how properties of concepts under the same parent relate to each other.
For example, a property being plausible for only one in a set of siblings
makes this property remarkable for the one concept.\\

\subsection{Grounding of Dependencies}
\label{sec:grounding}

The specified first-order constraints need to be grounded with the candidate statements
in a CSK collection, yielding a set of logical clauses (i.e., disjunctions of positive or negated
atomic statements).
To avoid producing a huge amount of clauses, 
we restrict the grounding to 
existing subject-property pairs and the high-confidence (>0.4) relationships of the WebIsALOD taxonomy
(avoiding its noisy long tail).

\paragraph{Expansion to similar properties}
Following this specification, the clauses would apply only for the same property of inter-related concepts,
for example, {\em eats meat} for {\em lions, leopards, hyenas} etc.
However, the CSK candidates may express the same or very similar properties in
different ways: {\em lions: eat meat}, {\em leopards: are carnivores}, {\em hyenas: eat carcasses} etc.
Then the grounded formulas would never trigger any inference, as the $p$ values are different.
%
We solve this issue by considering the similarity of different $p$ values based on
word-level embeddings (see Section \ref{sec:priorscores}).
For each property pair $(p_1, p_2)\in\mathcal{P}^2$, 
grounded clauses are generated if $\text{sim}(p_1, p_2)$ exceeds a threshold $t$.

We consider such highly related property pairs also for each concept alone, so that
we can deduce additional CSK statements by generating the following clauses:
$\forall s\in\mathcal{S}, \forall (p, q)\in\mathcal{P}^2,$
\begin{subequations}
\begin{align}
    \text{sim}(p, q) \geq t \Rightarrow &\left(\text{Plausible}(s, p) \Leftrightarrow \text{Plausible}(s, q)\right), \label{eq:similarity_plausible}\\
    &  \left(\text{Typical}(s, p) \Leftrightarrow \text{Typical}(s, q)\right),\\
    &  \left(\text{Remarkable}(s, p) \Leftrightarrow \text{Remarkable}(s, q)\right),\\
    &  \left(\text{Salient}(s, p) \Leftrightarrow \text{Salient}(s, q)\right)
\end{align}
\label{eq:similarity_contraints}
\end{subequations}
This expansion of the reasoning machinery allows 
us to deal with the noise and sparsity in the pre-existing CSK collections.

\paragraph{Weighting clauses}
Each of the atomic statements $\text{Plausible}(s,p)$, $\text{Typical}(s,p)$, $\text{Remarkable}(s,p)$ and $\text{Salient}(s,p)$
has a prior weight based on the confidence score from the underlying collection of CSK candidates
(see Sec.~\ref{sec:priorscores}).
These priors are denoted $\pi(s, p), \tau(s, p), \rho(s, p),$ and $\sigma(s, p)$.

Each grounded clause $c$ has three different weights:
\begin{enumerate}
    \item $\omega_r$, the weight of the logical dependency from which the clause is generated,
    a hyper-parameter for tuning the relative influence of different kinds of dependencies.
    \item $\omega_s$, the similarity weight, $\text{sim}(p_1, p_2)$ for clauses resulting from similarity expansion, or 1.0 if concerning only 
    a single property.
    \item $\omega_e$, the evidence weight, computed by combining the statistical priors
    for the individual atoms of the clause, using basic probability calculations for
    logical operators: $1-u$ for negation and $u+v - uv$ for disjunction with weights
    $u,v$ for the atoms in a clause.
\end{enumerate}
The final weight of a clause $c$ is computed as:
$$\omega^c = \omega_r\omega_s\omega_e$$
Table \ref{tab:clauses} shows a few illustrative examples.

\begin{table*}
    \centering
    \begin{tabular}{ll|ccc|c}
    \toprule
    \textbf{Rule} & \textbf{Clause} & $\omega_r$ & $\omega_s$ & $\omega_e$ & $\omega^c$ \\ \midrule
    \ref{eq:typical_implies_plausible} & Plausible(\textit{car}, \textit{hit wall}) $\vee$ $\neg$ Typical(\textit{car}, \textit{hit wall}) & 0.48 & 1 & 0.60 & 0.29\\
    \ref{eq:similarity_plausible} & Plausible(\textit{bicycle}, \textit{be at city}) $\vee$ $\neg$ Plausible(\textit{bicycle}, \textit{be at town}) & 0.85 & 0.86 & 1 & 0.73 \\
    \ref{eq:similarity_plausible} & Plausible(\textit{bicycle}, \textit{be at town}) $\vee$ $\neg$ Plausible(\textit{bicycle}, \textit{be at city}) & 0.85 & 0.86 & 1 & 0.73 \\
    \ref{eq:typical_prevents_remarkable_children} & $\neg$ Remarkable (\textit{bicycle}, \textit{transport person and thing}) $\vee$ $\neg$ Typical(\textit{car}, \textit{move person}) & 0.51 & 0.78 & 0.96 & 0.38 \\
    \bottomrule
    \end{tabular}
    \caption{Examples of grounded clauses with their weights (based on ConceptNet).}
\label{tab:clauses}
\end{table*}


\subsection{Integer Linear Program}

\paragraph{Notations} 
For reasoning over the validity of candidate statements, for each of the four facets,
we view every candidate statement 
$\textit{Facet}(s, p)$ 
as a variable  $v\in\mathcal{V}$, and its prior (either $\tau$, $\pi$, $\rho$ or $\sigma$, 
see Section \ref{sec:priorscores}) is denoted as $\omega^v$. 
Every grounded clause $c\in\mathcal{C}$, normalized into a disjunctive formula, 
can be split into 
variables with positive polarity, $c^+$,
and variables with negative polarity, $c^-$.

By viewing all $v$ as Boolean variables, 
we can now interpret the reasoning task as a 
weighted maximum satisfiability (Max-Sat) problem: 
find a truth-value assignment to the variables $v\in\mathcal{V}$
such that the sum of weights of satisfied clauses is maximized.
This is a classical NP-hard problem, but the literature offers a wealth of
approximation algorithms (see, e.g., \cite{manquinho2009algorithms}).
Alternatively and preferably for our approach, we can re-cast the Max-Sat problem into
a problem for integer linear programming (ILP) \cite{vazirani2013approximation}
where the variables $v$  become 0-1 decision variables.
Although ILP is more general and potentially more expensive than Max-Sat,
there are highly optimized and excellently engineered methods available in
software libraries like Gurobi \cite{gurobi}.
Moreover, we are ultimately interested not just in computing accepted variables (set to 1)
versus rejected ones (set to 0), but want to obtain an informative ranking of 
the candidate statements. To this end, we can relax an ILP into a fractional LP
(linear program), based on principled foundations \cite{vazirani2013approximation}, as discussed below.
Therefore, we adopt an ILP approach, with the following objective function and constraints:


\begin{equation}
    \max \sum_{v\in\mathcal{V}}\omega^vv + \sum_{c\in\mathcal{C}}\omega^cc
\end{equation}
under the constraints:
\begin{subequations}
\begin{align}
\forall c \in \mathcal{C} \>\> \forall v \in c^+ \>\> c - v & \geq 0 \label{constraint_pos}\\
\forall c \in \mathcal{C} \>\> \forall w \in c^- \>\> c + w -1 & \geq 0 \label{constraint_neg}\\
\forall c \in \mathcal{C} \>\> \sum_{v\in c^+} v + \sum_{w\in c^-} (1-w) - c & \geq 0 \label{constraint_imp}\\
\forall v \in \mathcal{V} \>\> v & \in [0, 1] \label{eq:constraint_v}\\
\forall c \in \mathcal{C} \>\> c & \in [0, 1] \label{eq:constraint_c}
\end{align}
\end{subequations}
Each clause $c$ is represented as a triple of ILP constraints, where Boolean operations $\neg$ and $\vee$ are
encoded via inequalities.

\subsection{Ranking of CSK Statements}

The ILP returns 0-1 values for the decision variables; so we can only accept or reject a
candidate statement.
Relaxing the ILP into an ordinary linear program (LP) drops the
integrality constraints on the decision variables, and would then
return fractional values for the variables. Solving an LP is typically
faster than solving an ILP.

The fractional values returned by the LP are not easily interpretable.
We could employ the method of randomized rounding \cite{DBLP:journals/combinatorica/RaghavanT87}: for fractional value
$x\in[0,1]$ we toss a coin that shows 1 with probability $x$ and 0 with
probability $1-x$.
This has been proven to be a constant-factor approximation (i.e., near-optimal solution)
on expectation.

However, we are actually interested in using the relaxed LP to compute
principled and informative rankings for the candidate statements.
To this end, we leverage the theory of {\em reduced costs}, aka. {\em opportunity costs}
\cite{bertsimas1997introduction}.
For an LP of the form {\em minimize $c^T x$ subject to $Ax \le b$ and $x \ge b$} with
coefficient vectors $c,b$ and coefficient matrix $A$, the reduced cost of variable $x_i$
that is zero in the optimal solution is the amount by which the coefficient
$c_i$ needs to be reduced in order to yield an optimal solution with $x_i > 0$.
This can be computed for all $x$ as $c - A^Ty$.
For maximization problems, the reduced cost is an increase of $c$.
Modern optimization tools like Gurobi directly yield these measures of sensitivity
as part of their LP solving. 

We use the reduced costs of the $x_i$ variables as a principled way of ranking them;
lowest cost ranking highest (as their weights would have to be changed most to
make them positive in the optimal solution).\\
As all variables with reduced cost zero would have the same rank,
we use the actual variable values (as a cue for the corresponding statement or dependency
being satisfied) as a tie-breaker.






\subsection{Scalability}



LP solvers are not straightforward to scale to cope with large amounts of input data.
For reasoning over all candidate statements in one shot, we would have to solve an LP with millions of variables. 
We devised and utilized the following technique to overcome this bottleneck in our experiments.

The key idea is to consider only limited-size neighborhoods in the taxonomic hierarchy
in order to partition the input data. 
In our implementation, to reason about the facets for a candidate statement $(s,p)$,
we identify the parents and siblings of $s$ in the taxonomy and then compile all
candidate statements and grounded clauses where at least one of these concepts appears.
This typically yields subsets of size in the hundreds or few thousands.
Each of these forms a partition, and we generate and solve an LP for each partition separately.
This way, we can run the LP solver on many partitions independently in parallel.
The partitions overlap, but each $(s,p)$ is associated with a primary partition
with the statement's specific neighborhood.



\section{Prior Statistics}
\label{sec:priorscores}

So far, we assumed that prior scores
-- $\pi(s,p), \tau(s,p), \rho(s,p), \sigma(s,p)$ --
are given, in order to compute weights for
the ILP or LP.
This section explains how we obtain these priors.
In a nutshell, we obtain basic scores
from the underlying CSK collections and their
combination with embedding-based similarity, and from textual entailment and relatedness in
the taxonomy (Subsection \ref{sec:priorscores-basic}).
We then define aggregation functions
to combine these various cues
(Subsection \ref{sec:priorscores-aggr}).

\subsection{Basic Scores}
\label{sec:priorscores-basic}

Basic statements like $(s,p)$ are taken from
existing CSK collections, which often provide
{\em confidence scores} based on observation
frequencies or human assessment (of 
crowdsourced statements or samples).
We combine these confidence measures,
denoted $\text{score}(s,p)$ with embedding-based
similarity between two properties, 
$\text{sim}(p,q)$.
%
%
%
Each property $p$ is tokenized into a bag-of-words $\{w_1, \dots, w_n\}$ and encoded as the 
idf-weighted centroid of the embedding vectors $\vec{w_i}$ obtained from a pre-trained word2vec
model\footnote{\small\url{https://code.google.com/archive/p/word2vec/}{GoogleNews-vectors-negative300.bin.gz}}:
$\vec{p} = \sum_{i=1}^n \text{idf}(w_i) \> \vec{w}_i$.
The similarity between two properties is the cosine between the vectors  mapped into $[0,1]$:
$\text{sim}(p, q) = \frac{1}{2}\left(\frac{\langle\vec{p}, \vec{q} \rangle}{\left\lVert \vec{p} \right\rVert \left\lVert \vec{q} \right\rVert}+1\right)$.



Confidence scores and similarities are
then combined and normalized into
a quasi-probability:
$$\mathbb{P}[s, p] = \frac{1}{Z} \sum_{\substack{q\in\mathcal{P} \\ \text{sim}(p, q)\geq t}}  \text{score}(s, q) \times \text{sim}(q, p) $$
where $Z$ is a normalization factor and $t$ 
is a threshold (set to 0.75 in our implementation).
The intuition for this measure is
that it reflects the probability of
$(s,p)$ 
being observed in the digital world,
where evidence is
accumulated over different phrases for
inter-related properties such as
{\em eat meat}, {\em are carnivores}, 
{\em are predators}, {\em prey on antelopes} etc.\\

We can now derive additional measures
that serve as building blocks for the
final priors:
\begin{itemize}
    \item the marginals $\mathbb{P}[s]$ for subjects and $\mathbb{P}[p]$ for properties,
    \item the conditional probabilities of observing $p$ given $s$, or the reverse; $\mathbb{P}[p \mid s]$ can be thought of as the \emph{necessity} of the property $p$ for the subject $s$, while $\mathbb{P}[s \mid p]$ can be thought of as a \emph{sufficiency} measure,
    \item the probability that the observation of $s$ implies the observation of $p$, which can be expressed as:
    $$
    \begin{array}{rcl}
        \mathbb{P}[s \Rightarrow p] 
         & = & 1 - \mathbb{P}[s] + \mathbb{P}[s, p] \\
    \end{array}
    $$
\end{itemize}

Beyond aggregated frequency scores, priors rely on two more components, scores from textual entailment models and taxonomy-based information gain.

\noindent{\bf Textual entailment:}\\
A variant of $P[s \Rightarrow p]$
is to tap into corpora and learned models
for textual entailment:
does a sentence such as ``Simba is a lion''
entail a sentence ``Simba lives in a pride''?
%
%
%
%
We leverage the
attention model from the AllenNLP project \cite{gardner-etal-2018-allennlp}
learned from the SNLI corpus \cite{snli:emnlp2015} and other annotated text collections.
This gives us scores for two measures:
does $s$ entail $p$, $\text{entail}(s \rightarrow p)$,
and does $p$ contradict $s$, $\text{con}(s, p)$.

\vspace{0.2cm}
\noindent{\bf Taxonomy-based information gain:}\\
For each $(s,p)$ we define a neighborhood of concepts, $N(s)$, by the parents and siblings of $s$,
and consider all statements for $s$ versus all statements for $N(s)-\{s\}$ as a potential cue
for remarkability.
For each property $p$ and concept set $S$, the entropy of $p$ is 
$H(p|S) = \frac{1}{X_S} \log{X_S} + \frac{X_S-1}{X_S} \log{\frac{X_S}{X_S-1}}$
where $X_S = |\{q~|~\exists s\in S: (s,q)\}|$.
Instead of merely count-based entropy, we could also incorporate relative weights of different properties,
but the as a basic cue, the simple measure is sufficient.
Then, the information gain of $(s,p)$ is $IG(s,p) = H(p\mid\{s\})$ $- H(p\mid S-\{s\})$.

\subsection{Score Aggregation}
\label{sec:priorscores-aggr}

All the basic scores -- $\mathbb{P}[s,p]$, $\mathbb{P}[s\mid p]$, $\mathbb{P}[p\mid s]$, 
$\mathbb{P}[s \Rightarrow p]$, $\text{entail}(s \rightarrow p)$, $\text{con}(s,p)$
and $IG(s,p)$ -- 
are fed into regression models that learn an aggregate score
for each of the four facets: plausibility, typicality, remarkability and
saliency.
The regression parameters (i.e., weights for the different basic scores)
are learned from small set of facet-annotated CSK statements,
separately, for each of the four facets.
We denote the aggregated scores, serving as priors for the reasoning step,
as $\pi(s,p)$, $\tau(s,p)$, $\rho(s,p)$ and $\sigma(s,p)$.

\section{Experiments}

We evaluate three aspects of the \dice{} framework: (i) accuracy
in ranking statements along the four CSK facets, 
(ii) run-time and scalability,
(iii) the ability to enrich CSK collections with 
newly inferred statements.
The main hypothesis under test is how well \dice{}
can rank statements for each of the four CSK facets.
We evaluate this by obtaining crowdsourced judgements
for a pool of sample statements.

\subsection{Setup}

\paragraph{Datasets}
We use three CSK collections for evaluating the added value
that \dice{} provides: (i) ConceptNet, a crowdsourced, sometimes wordy collection of general-world CSK. (ii) Tuple-KB, a CSK collection extracted from web sources
with focus on the science domain, with comparably short and canonicalized SPO triples. (iii) Quasimodo, a web-extracted general-world CSK collection with 
focus on saliency. 
Statistics on these datasets are shown in Table~\ref{tab:input}.

\begin{table}
    \centering
    \begin{tabular}{l|cc}
        \toprule
        \textbf{CSK collection} & \textbf{\#subjects} & \textbf{\#statements} \\ \midrule
        Quasimodo & 13,387 & 1,219,526  \\
        ConceptNet & 45,603 & 223,013  \\
        TupleKB & 28,078 & 282,594 \\
        \bottomrule
    \end{tabular}
    \caption{Input CSK collections.}
\label{tab:input}
\end{table}

\begin{table}
    \centering
    \scalebox{0.95}{
    \begin{tabular}{l|ccc}
        \toprule
        \textbf{CSK collection} & \textbf{\#nodes} & \textbf{\#parents/node} & \textbf{\#siblings/node}\\ \midrule
        Quasimodo & 11148 & 15.33 & 3627.8 \\
        ConceptNet & 41451 & 1.15 & 63.7 \\
        TupleKB & 26100 & 2.14 & 105.1 \\
	    Music-manual & 8 & 1.68 & 3.4 \\
        \bottomrule
    \end{tabular}
    }
    \caption{Taxonomy statistics.} 
\label{tab:taxonomies}
\end{table}

\begin{table*}
   \centering
   \scalebox{0.95}{
    \begin{tabular}{r|c|cc|cc|cc|cc}
    \toprule
    \multirow{2}{*}{\textbf{Dimension}} & \multirow{2}{*}{\textbf{Random}} & \multicolumn{2}{c|}{\textbf{ConceptNet}} & \multicolumn{2}{c|}{\textbf{TupleKB}} & \multicolumn{2}{c|}{\textbf{Quasimodo}} & \multicolumn{2}{c}{\textbf{Music-manual}} \\
    &  & \textbf{Baseline~\cite{conceptnet}} & \textbf{\dice} & \textbf{Baseline~\cite{TupleKB}} & \textbf{\dice} & \textbf{Baseline~\cite{quasimodo}} & \textbf{\dice} & \textbf{Baseline~\cite{conceptnet}} & \textbf{\dice} \\
    \midrule
    Plausible  & 0.5 & 0.52 & 0.62 & 0.53 & 0.57 & 0.57 & 0.59 & 0.21 & \textbf{0.67} \\
    Typical    & 0.5 & 0.39 & \textbf{0.65} & 0.37 & 0.59 & 0.52 & \textbf{0.64} & 0.54 & \textbf{0.70} \\
    Remarkable & 0.5 & 0.52 & \textbf{0.69} & 0.50 & 0.54 & 0.56 & 0.56 & 0.49 & \textbf{0.74} \\
    Salient    & 0.5 & 0.54 & 0.65 & 0.59 & 0.61 & 0.53 & 0.63 & 0.51 & 0.65 \\
    \midrule
    Avg.    & 0.5 & 0.50 & \textbf{0.66} & 0.50 & 0.58 & 0.54 & 0.61 & 0.52 & \textbf{0.69} \\
    \bottomrule
    \end{tabular}
    }
    \caption{Precision of pairwise preference (ppref) of \dice{}\ versus original CSK collections. Significant gains over baselines ($\alpha=0.05)$ are boldfaced. }
\label{tab:mainresults}
\end{table*}

To construct taxonomies for each of these collections, we utilized the WebIsALOD dataset~\cite{hertling2017webisalod}, a web-extracted noisy set of ranked subsumption pairs (e.g., {\tt{\small tiger isA big\_cat}} - 0.88, {\tt{\small tiger isA carnivore}} - 0.83). 
We prune out long-tail noise by setting a threshold of 0.4 for
the confidence scores that WebIsALOD comes with. 
To evaluate the influence of taxonomy quality, we also hand-crafted 
a small high-quality taxonomy for the music domain, 
with 10 concepts and 9 subsumption pairs, such as \texttt{\small rapper} being a subclass of \texttt{\small singer}.
%
Table \ref{tab:taxonomies} gives statistics on the taxonomies 
per CSK collection.
Differences between \#nodes in Table \ref{tab:taxonomies} 
and \#subjects in Table~\ref{tab:input} are 
caused by merging nodes on hypernymy paths without branches (\#children=1).

\paragraph{Annotation}
To obtain labelled data for hyper-parameter tuning and 
as ground-truth for evaluation, we conducted a crowdsourcing project
using Amazon Mechanical Turk.
For saliency, typicality and remarkability,  we sampled 200 subjects each with 2 properties from each of the CSK collections, and asked annotators for pairwise preference with regard to each of the three facets, using a 5-point Likert scale.
That is, we show two statements for the same subject, and the annotator
could slide on the scale between 1 and 5 to indicate the more 
salient/typical/remarkable statement.
For the plausibility dimension, we sampled 200 subjects each with two properties, and asked annotators to 
assess the plausibility of individual statements on a 5-point scale. 
Then we paired up two statements for the same subject as a 
post-hoc preference pair.
The rationale for this procedure is to avoid biasing the annotator
in judging plausibility by showing two statements at once,
whereas it is natural to compare pairs on the other three dimensions.

\begin{figure}
    \centering
     \includegraphics[scale=.62]{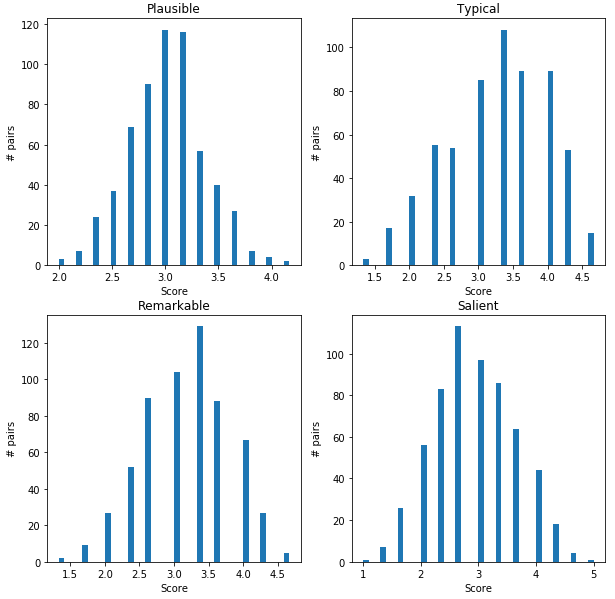}
     \caption{Aggregate label distribution.}
     \label{fig:labeldistribution}
 \end{figure}

In total, we had $4\times 4 \times200=3200$ tasks, each given to 3 annotators. 
The final scores for each statement and facet were the 
averages of the three numerical judgments. 
Regarding inter-annotator agreement, we observed a 
reasonably low standard deviation of 0.81/0.92/0.98/0.92 
(over the scale from 1 to 5) for the dimensions plausibility/\-typicality/\-remarkability/\-saliency on ConceptNet, with similar values on the other CSK collections. Aggregate label distributions are shown in Fig.~\ref{fig:labeldistribution}. When removing indeterminate 
samples, with avg.\ score between 2.5 and 3.5, and interpreting annotator scores as binary preferences, inter-annotator agreement was fair to moderate,
with Fleiss' Kappa values of 0.31, 0.30, 0.25 and 0.48
for plausibility, typicality, remarkability and saliency, respectively.

\paragraph{Evaluation Metrics}
In the actual evaluation, we used withheld pairwise annotations for statements along the dimensions plausibility, typicality, remarkability and
saliency as ground truth, and compared, for each system score, for how many of these pairs its scores implicated the same ordering, i.e., measured the \emph{precision in pairwise preference} (ppref)~\cite{ppref}.

\paragraph{Hyper-parameter tuning}
The 800 labeled statements per CSK collection were split into 70\% for hyper-parameter optimization and 30\% for evaluation. 
We performed two hyper-parameter optimization steps.
In step 1, we learned the weights for aggregating the basic scores by a regression model based on interpreting pairwise data as single labels (i.e., the preferred property is labelled as 1, the other one as 0). 
In step 2, we used Bayesian optimization to tune the weights of the 
constraints. As exhaustive search was not possible, we used the Tree-structured Parzen Estimator (TPE) algorithm from the Hyperopt \cite{bergstra2013making} library. We used the 0-1 loss function on the ordering of the pairs as metric,
and explored the search space in two ways: 
\begin{enumerate}
    \item discrete exploration space $\{0, 0.1, 0.5, 1\}$, followed by
    \item continuous exploration space of radius 0.2 centered on the value selected in the previous step.
\end{enumerate}
%
For ConceptNet, constraints were assigned an average weight of 0.404, with the highest weights for: (\ref{eq:similarity_contraints}) Similarity constraints (weight 0.85), (\ref{eq:plausibility_inference_constraint}) Plausibility inference (weight 0.66) and (\ref{eq:not_plausible_siblings_contraints}) Sibling implausibility implying remarkability (weight 0.60).
All constraints were assigned non-negligible positive weights; so they are all important for joint inference.



\begin{table}
\centering
   \scalebox{0.95}{
\begin{tabular}{@{}r|cc|c@{}}
\toprule
 & \begin{tabular}[c]{@{}c@{}}\textbf{Priors}\\ \textbf{only}\end{tabular} & \multicolumn{1}{c|}{\begin{tabular}[c]{@{}c@{}}\textbf{Constraints} \\ \textbf{only}\end{tabular}} & \textbf{Both} \\ \midrule
Plausible & 0.54 & 0.51 & 0.62 \\
Typical & 0.53 & 0.42 & 0.65 \\
Remarkable & 0.65 & 0.57 & 0.69 \\
Salient & 0.56 & 0.52 & 0.65 \\ \midrule
Avg. & 0.58 & 0.51 & 0.66 \\
\bottomrule
\end{tabular}
}
\caption{Ablation study using ConceptNet as input.}
\label{tab:ablation}
\end{table}

\begin{table}
\centering
   \scalebox{0.95}{
\begin{tabular}{@{}c|cccc@{}}
\toprule
\bfseries Ranking & \bfseries Existing & \multicolumn{3}{c}{\bfseries New statements} \\
\bfseries dimension & \bfseries statements & \bfseries 25\% & \bfseries 50\% & \bfseries 100\% \\
\midrule
Plausible & \multirow{2}{*}{3.44} & 3.54 & 3.43 & 3.41  \\
Typical &  & 3.27 & 3.31 & 3.26  \\
\bottomrule
\end{tabular}
}
\caption{Plausibility of top-ranked newly inferred statements with ConceptNet as input.}
\label{tab:enrichment}
\end{table}

\begin{table}
\centering
\begin{tabular}{@{}ll@{}}
\toprule
\textbf{Subject} & \textbf{Novel properties} \\ \midrule
sculpture & be at art museum, be silver or gold in color \\
athlete & requires be good sport, be happy when they win \\
saddle & be used to ride horse, be set on table \\ \bottomrule
\end{tabular}
\caption{Examples of new statements inferred by \dice\ with ConceptNet as input.}
\label{tab:anecdotalenrichment}
\end{table}

\begin{table*}
\centering
\scalebox{0.95}{
\begin{tabular}{@{}lllllll@{}}
\toprule
\multirow{2}{*}{\textbf{Subject}} & \multicolumn{1}{c}{\multirow{2}{*}{\textbf{Property}}} & \multicolumn{1}{c}{\textbf{Baseline}} & \multicolumn{4}{c}{\textbf{\dice}} \\
 & \multicolumn{1}{c}{} & \textbf{CN-score} & \textbf{plausible} & \textbf{typical} & \textbf{remarkable} & \textbf{salient} \\ \midrule
snake & be at shed & 0.46 & 0.29 & 0.71 & 0.29 & 0.18 \\
snake & be at pet zoo & 0.46 & 0.15 & 0.29 & 0.82 & 0.48 \\
snake & bite & 0.92 & 0.58 & 0.13 & 0.61 & 0.72 \\
lawyer & study legal precedent & 0.46 & 0.25 & 0.73 & 0.37 & 0.18 \\
lawyer & prove that person be guilty & 0.46 & 0.06 & 0.47 & 0.65 & 0.40 \\
lawyer & present case & 0.46 & 0.69 & 0.06 & 0.79 & 0.75 \\
bicycle & requires coordination & 0.67 & 0.62 & 0.40 & 0.36 & 0.35 \\
bicycle & be used to travel quite long distance & 0.46 & 0.30 & 0.20 & 0.77 & 0.64 \\
bicycle & be power by person & 0.67 & 0.19 & 0.33 & 0.66 & 0.55 \\ \bottomrule
\end{tabular}
}
\caption{Anecdotal examples from \dice{} run on ConceptNet.}
\label{tab:anecdotal_examples}
\end{table*}


\subsection{Results}

\paragraph{Quality of rankings}
Table~\ref{tab:mainresults} shows the main result of our experiments:
the precision in pairwise preference (ppref) scores~\cite{ppref},
that is, the fraction of pairs where \dice\ or a baseline produced the same ordering as the crowdsourced ground-truth. 
As baseline,
we rank all statements by the confidence scores from the original CSK collections, 
which implies that the ranking is identical for all four dimensions. 
As the table shows, \dice\ consistently outperforms the baselines by a large margin of 7 to 18 percentage points. 
It is also notable that scores in the original ConceptNet and TupleKB are
negatively correlated with typicality (values lower than 0.5), pointing out a substantial fraction
of valid but not exactly typical properties in these pre-existing CSK collections.

\paragraph{Ablation study}
To study the impact of statistical priors and constraint-based reasoning, 
we compare two variants of \dice{}: (i)
using only priors without the reasoning stage, and 
(ii) using only the constraint-based reasoning with all priors set to 0.5. 
The resulting ppref scores are shown in Table~\ref{tab:ablation}. 
In isolation, priors and reasoning perform 8 and 15 percentage points worse than the combined \dice{} method.
This clearly demonstrates the importance of both stages and the synergistic benefit from their interplay.

\paragraph{Enrichment potential}
All CSK collections are limited in their coverage of long-tail concepts. 
By exploiting the taxonomic and embedding-based relatedness between different concepts,
we can generate candidate statements that were not observed before (e.g., because online contents
rarely talk about generalized concepts like big cats, and mostly mention only properties of lions, leopards,
tigers etc.).
As mentioned in Section~\ref{sec:grounding},
simple templates can be used to generate candidates.
These are fed into \dice{} reasoning together with the statements that are actually contained in the
existing CSK collections.

To evaluate the quality of the \dice{} output for such ``unobserved'' statements, 
we randomly sampled 10 ConceptNet subjects, and grounded the reasoning framework for these subjects for all properties observed in their taxonomic neighbourhood (i.e., parents and siblings). 
We then asked annotators to assess the plausibility of 100 sampled statements.

To compute the quality of \dice{} scores, we consider the top-ranked statements by predicted plausibility 
and by typicality, where we
vary the recall level: number of statements from the ranking in relation to the number of statements
that ConceptNet contains for the sampled subjects.
The results are shown in Table~\ref{tab:enrichment} for recall 25\%, 50\%  and 100\%,
that is up to doubling the size of ConceptNet for the given subjects.
As one can see, \dice\ can expand the pre-existing CSK by 25\% without losing in quality,
and even up to 100\% expansion the decrease in quality is negligible.
Table~\ref{tab:anecdotalenrichment} presents anecdotal statements absent in ConceptNet.


\paragraph{Run-Time}
All experiments were run on a cluster with 40 cores and 500 GB memory.
Hyper-parameter optimization took 10-14 hours for each of the three CSK inputs. 
Computing the four-dimen\-sional scores for all statements took about 3 hours,
3 hours and 24 hours for ConceptNet, TupleKB and Quasimodo, respectively.

The computationally most expensive steps are the semantic similarity computation and 
the LP solving.
For semantic similarity computation, a big handicap is
 the verbosity and hence diversity of the phrases for properties
 (e.g., ``live in the savannah'', ``roam in the savannah'', ``are seen in the
 African savannah'', ``can be found in Africa's grasslands'' etc.).
 We observed on average 1.55 statements per distinct property for ConceptNet,
 and 1.77 for Quasimodo. 
 Therefore, building the input matrix for the LP is very time-consuming.
For LP solving, the Gurobi algorithm
has polynomial run-time in the number of variables. 
However, we do have a huge number of variables. Empirically, we need to cope with
about \textit{\#constraints} $\times$ \textit{\#statements}$^{1.2}$
variables.

\paragraph{Anecdotal examples}
Table~\ref{tab:anecdotal_examples} gives a few anecdotal outputs with scores returned by \dice.
Note that the scores produced do not represent probabilities, but global ranks (i.e., we percentile-normalized the scores produced by \dice, as they have no inherent semantics other than ranks). For instance, \texttt{\small be at shed} was found to be much more typical than \texttt{\small be at pet zoo} for \texttt{\small snake}, while salience was the other way around. Note also the low variation in ConceptNet scores, i.e., in addition to being unidimensional, this low variance makes any ranking difficult.

\section{Discussion}


\paragraph{Experimental results}
The experiments showed that \dice{} can capture CSK along the four dimensions significantly better than the single-dimensional baselines. The ablation study highlighted that a combination of prior scoring and constraint-based joint reasoning is highly beneficial (0.66 average ppref vs.\ 0.58 and 0.51 of each step in isolation, see Table~\ref{tab:ablation}). Among the dimensions, we find that plausibility is the most difficult of the four dimensions
(see Table~\ref{tab:mainresults}). 
The learning of hyper-parameters shows that all constraints are useful
and contribute to the outcome of \dice{}, with similarity 
dependencies and plausibility inference having the strongest influence. 

Comparing the three CSK collections that we worked with, 
we observe that the crowdsourced ConceptNet is a priori cleaner and
hence easier to process than Quasimodo and TupleKB. Also, manually designed taxonomies gave \dice{} a performance bost of 0.03-0.11 in ppref over the noisy web extracted WebIsALOD taxonomies.

\paragraph{Task difficulty} 
Scoring commonsense statements by dimensions beyond confidence has never been attempted before, and a major challenge is to design appropriate and varied input signals towards specific dimensions. Our experiments showed that  
\dice{} can approximate the human-generated ground-truth rankings to a considerable degree (0.58-0.69 average ppref), although a gap remains (see Table~\ref{tab:mainresults}). We conjecture that in order to approximate human judgments even better, more and finer-grained input signals, for example about textual contexts of statements, are needed.



\paragraph{Enriched CSK data}
Along with this paper, we publish six datasets: the 3 CSK collections ConceptNet, TupleKB and Quasimodo enriched by \dice{} with score for the four CSK dimensions,
and \pagebreak additional inferred statements that expand the original CSK data by
about 50\%. The datasets can be downloaded from \url{https://tinyurl.com/y6hygoh8}.

\begin{figure*}
        \centering
        \begin{subfigure}[b]{0.475\textwidth}
            \centering
            \fbox{\includegraphics[width=\textwidth]{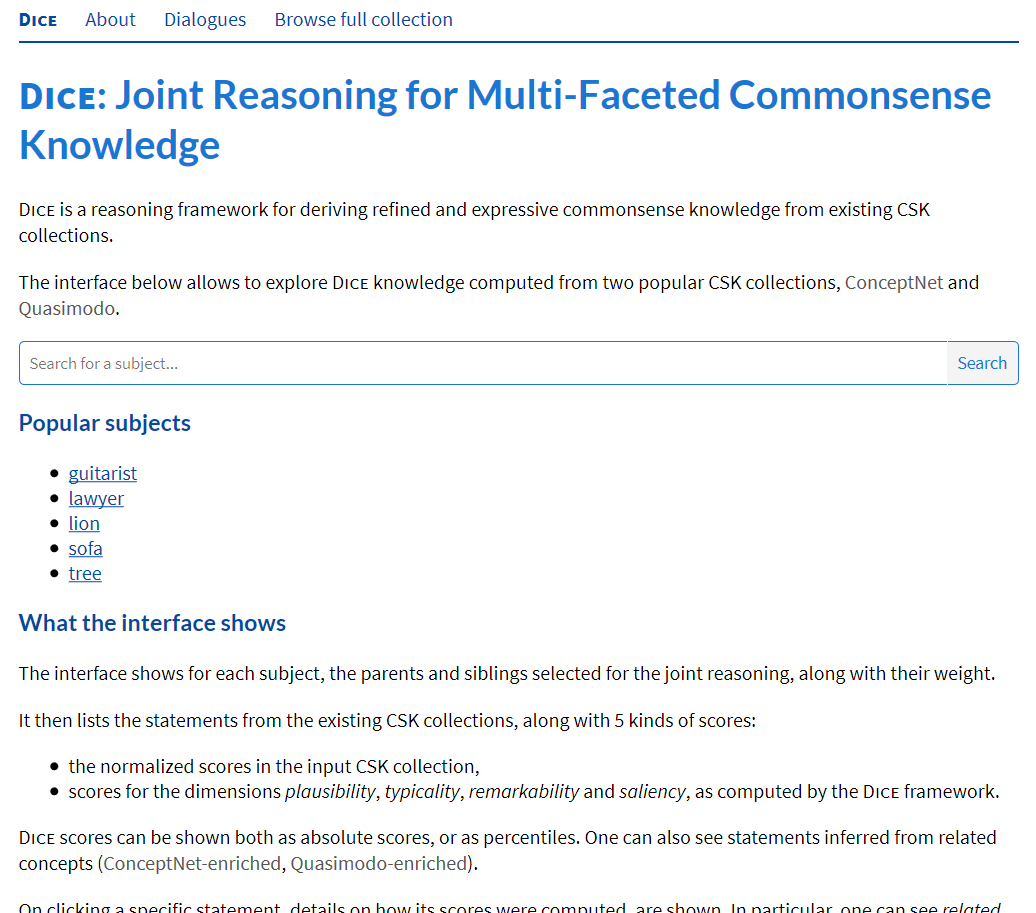}}
            \caption[unclear]%
            {{\small Demo landing page.}}    
        \end{subfigure}
        \hfill
        \begin{subfigure}[b]{0.475\textwidth}  
            \centering 
            \fbox{\includegraphics[width=\textwidth]{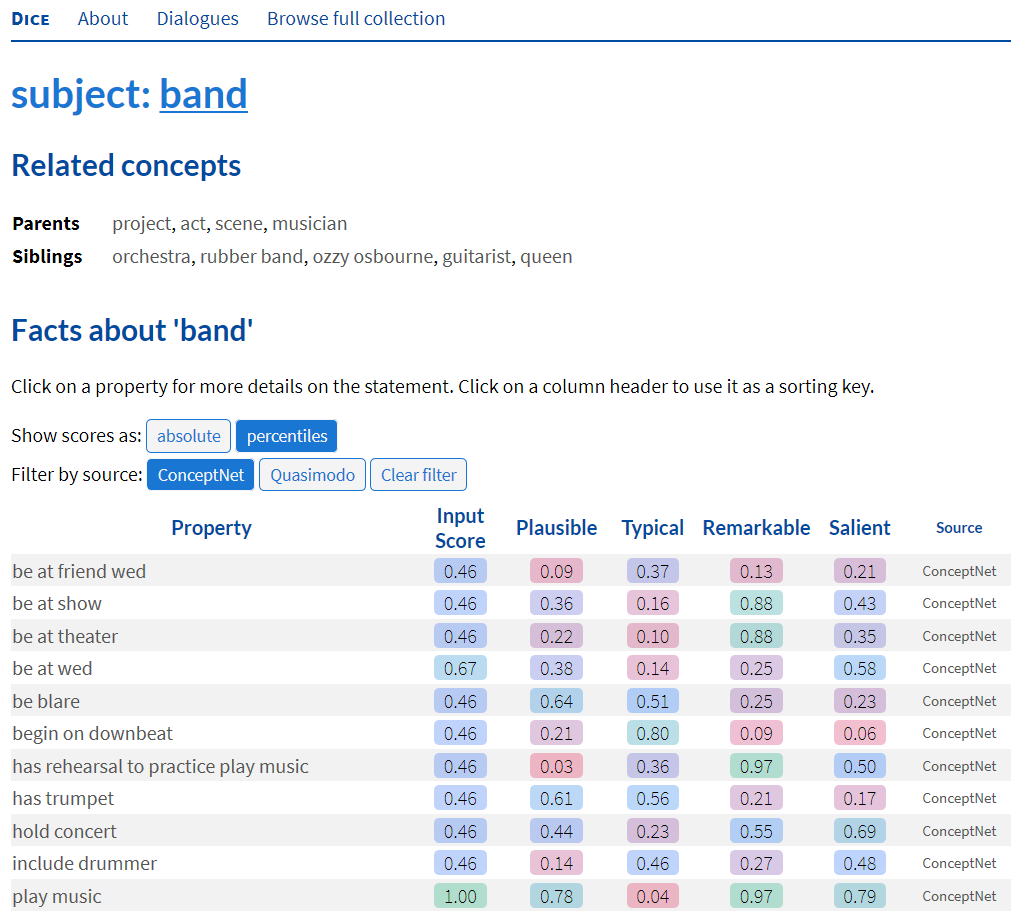}}
            \caption{{\small List of statements for subject \textit{band}.}}
        \end{subfigure}
        \vskip\baselineskip
        \begin{subfigure}[b]{0.475\textwidth}   
            \centering 
            \fbox{\includegraphics[width=\textwidth]{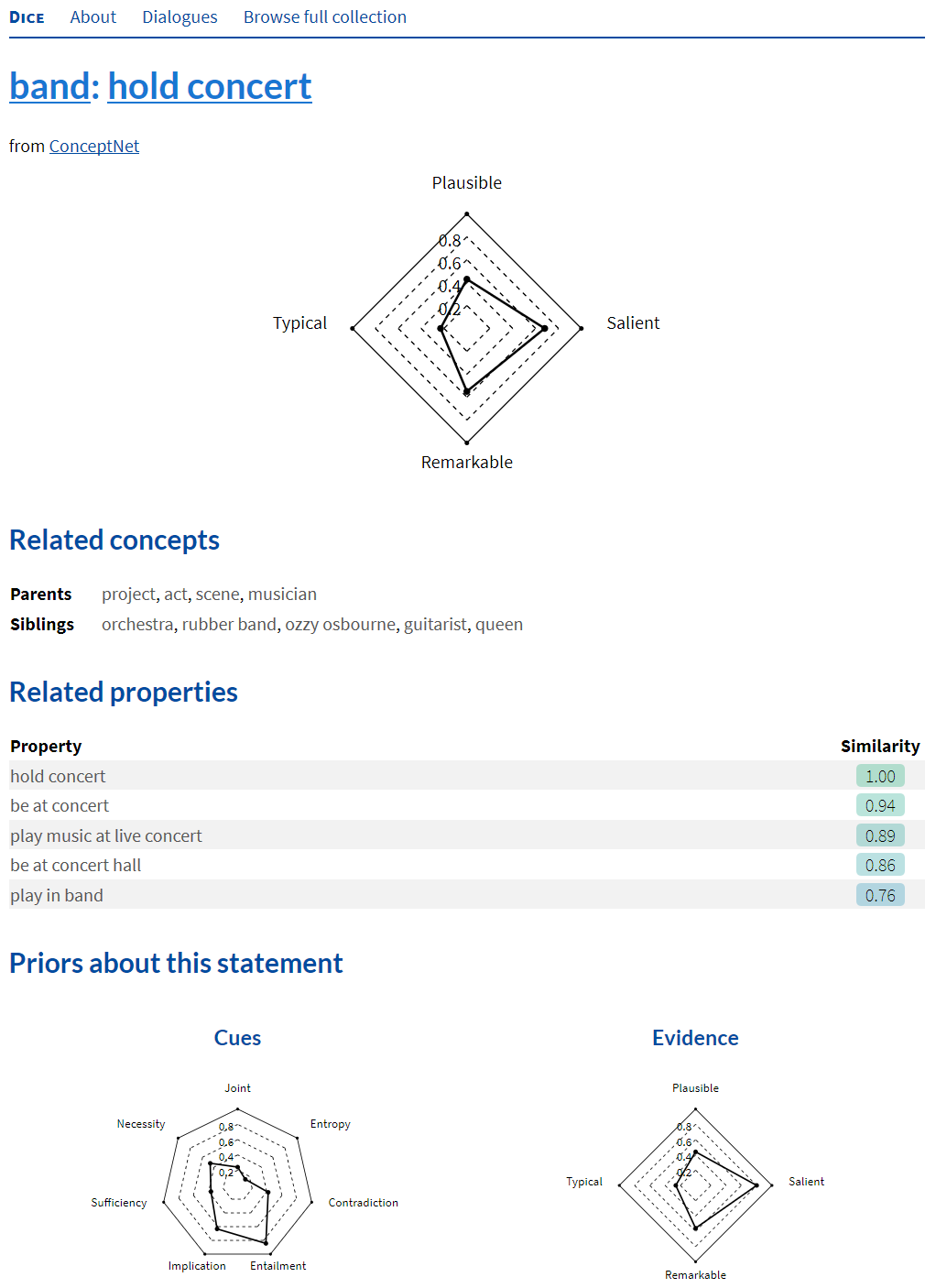}}
            \caption{{\small Scores and neighbourhood for statement \textit{band: hold concert}.}}    
        \end{subfigure}
        \quad
        \begin{subfigure}[b]{0.475\textwidth}   
            \centering 
            \fbox{\includegraphics[width=\textwidth,trim={0 15.3cm 0 0},clip]{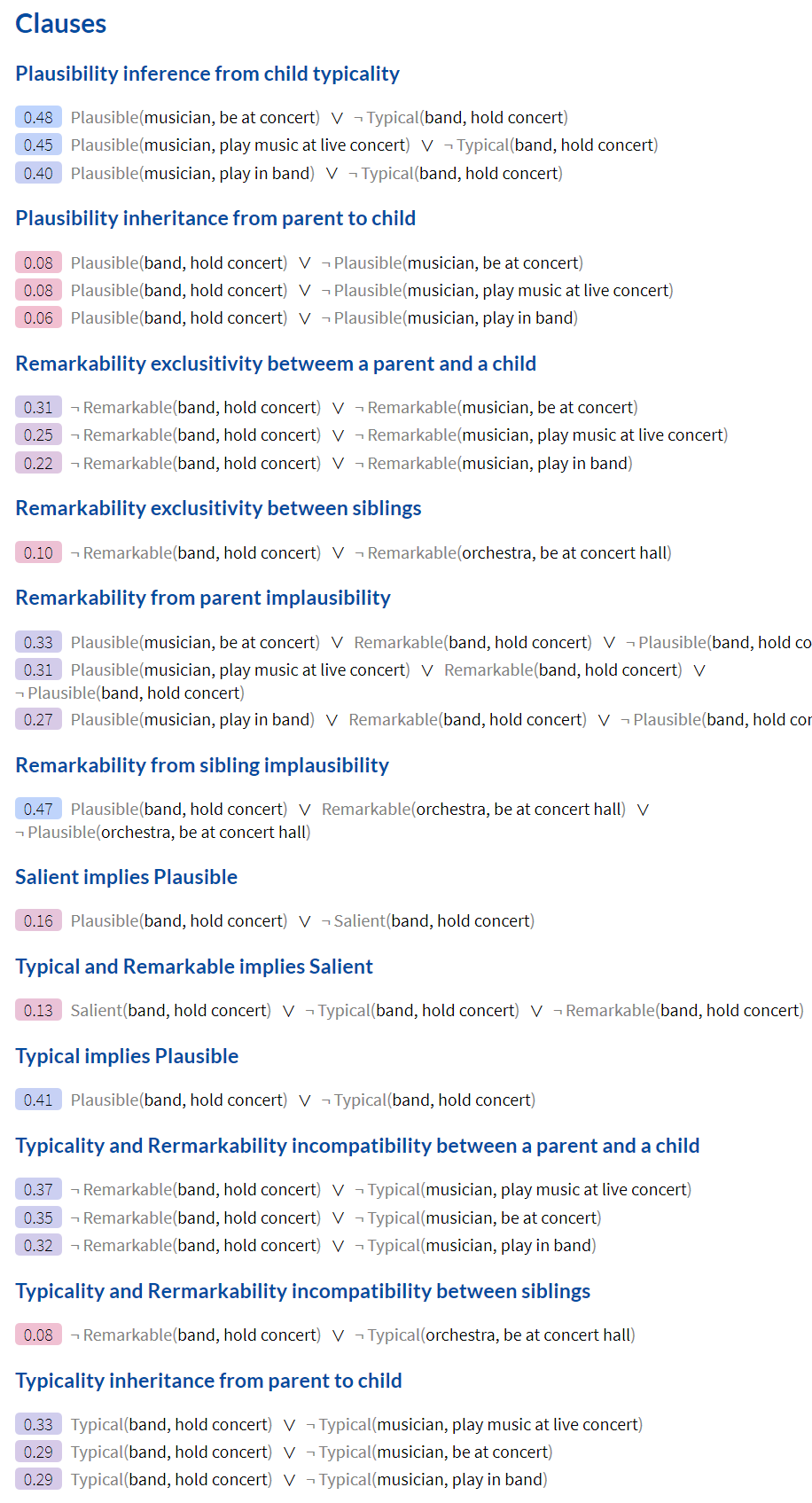}}
            \caption[]%
            {{\small Materialized clauses for statement \textit{band: hold concert}.}}    
            \label{fig:mean and std of net44}
        \end{subfigure}
        \caption{Screenshots from the web-based demonstration platform.} 
        \label{fig:demo}
    \end{figure*}

\paragraph{Web demonstrator}
The results of running \dice{} on ConceptNet and Quasimodo are showcased in an interactive web-based demo. The interface shows original scores from these CSK collections as well as the per-dimension scores computed by \dice. 
Users can explore the values of individual cues, the priors, the taxonomic neighborhood of a subject, and the clauses generated by the rule grounding. The demo is available online at \url{https://dice.mpi-inf.mpg.de}, we also show
 screenshots in Figure~\ref{fig:demo}.

From a landing page (Fig.~\ref{fig:demo}(a)), users can navigate to individual subjects like \emph{band} (Fig.~\ref{fig:demo}(b)). On pages for individual subjects, taxonomic parents and siblings are shown at the top, followed by commonsense statements from ConceptNet and Quasimodo. For each statement, its normalized score or percentile in its original CSK collection, along with scores and percentiles along the four dimensions as computed by \dice{}, are shown.
Colors from green to red highlight to which quartile a percentile value belongs.
On inspecting a specific statement, e.g., \emph{band: hold concert} (Fig.~\ref{fig:demo}(c)), one can see related statements used for computing basic scores, along with the values of the priors and evidence scores. Further down on the same page (Fig.~\ref{fig:demo}(d)), the corresponding materialized clauses from the ILP, along with their weight $\omega^c$, are shown.

\section{Conclusion}

This paper presented \dice, a joint reasoning framework for 
commonsense knowledge (CSK) that incorporates inter-dependencies between statements
by taxonomic relatedness and other cues.
This way we can capture more expressive meta-properties of concept-property statements
along the four dimensions of plausibility, typicality, remarkability and saliency.
This richer knowledge representation is a major advantage over prior works on
CSK collections.
In addition, we have devised techniques to compute informative rankings for
all four dimensions, using the theory of reduced costs for LP relaxation.
We believe that such multi-faceted rankings of CSK statements are crucial
for next-generation AI, particularly towards
more versatile and robust conversational bots.
Our future work plans include leveraging this rich CSK for advanced question answering
and human-machine dialogs.

\clearpage
\newpage

\balance

\bibliographystyle{myplain}
\bibliography{refs}

\end{document}